\documentclass[
twocolumn
]{ceurart}

\usepackage{listings}
\usepackage{graphicx}
\newcommand{\chunk}[2]{%
	\fcolorbox{black}{yellow}{\bfseries\sffamily\scriptsize#1}%
   {$\blacktriangleright$#2$\blacktriangleleft$}%
}
\newcommand{\ben}[1]{\chunk{BenjaminR}{\textbf{\textcolor{orange}{#1}}}}
\newcommand{\ale}[1]{\chunk{AlexanderF}{\textbf{\textcolor{blue}{#1}}}}

\begin{document}

%%
%% Rights management information.
%% CC-BY is default license.
\copyrightyear{2023}
\copyrightclause{Copyright for this paper by its authors.
  Use permitted under Creative Commons License Attribution 4.0
  International (CC BY 4.0).}

%%
%% This command is for the conference information
\conference{ConfWS'23: 25th International Workshop on Configuration, Sep 6--7, 2023, Málaga, Spain}

%%
%% The "title" command
\title{Solving Multi-Configuration Problems: A Performance Analysis with Choco Solver}

%\tnotemark[1]
%\tnotetext[1]{You can use this document as the template for preparing your
%  publication. We recommend using the latest version of the ceurart style.}

%%
%% The "author" command and its associated commands are used to define
%% the authors and their affiliations.
\author[1]{Benjamin Ritz}[%
orcid=0009-0000-7774-6693,
email=ritz@student.tugraz.at,
url=https://www.tugraz.at/,
]
\cormark[1]
%\fnmark[1]
%\address[1]{Peoples' Friendship University of Russia (RUDN University),
%  6 Miklukho-Maklaya St, Moscow, 117198, Russian Federation}
%\address[2]{Joint Institute for Nuclear Research,
%  6 Joliot-Curie, Dubna, Moscow region, 141980, Russian Federation}

\author[1]{Alexander Felfernig}[%
orcid=0000-0003-0108-3146,
email=alexander.felfernig@ist.tugraz.at,
url=https://www.felfernig.eu,
]

\author[1]{Viet-Man Le}[%
orcid=0000-0001-5778-975X,
email=vietman.le@ist.tugraz.at,
url=https://www.tugraz.at/,
]

\author[1]{Sebastian Lubos}[%
orcid=0000-0002-5024-3786,
email=vietman.le@ist.tugraz.at,
url=https://www.tugraz.at/,
]

%\fnmark[1]
\address[1]{Graz University of Technology, Inffeldgasse 16b, 8010 Graz, Austria}

%\author[4]{Manfred Jeusfeld}[%
%orcid=0000-0002-9421-8566,
%email=Manfred.Jeusfeld@acm.org,
%url=http://conceptbase.sourceforge.net/mjf/,
%]
%\fnmark[1]
%\address[4]{University of Skövde, Högskolevägen 1, 541 28 Skövde, Sweden}

%% Footnotes
\cortext[1]{Corresponding author.}
%\fntext[1]{These authors contributed equally.}

%%
%% The abstract is a short summary of the work to be presented in the
%% article.
\begin{abstract}
In many scenarios, configurators support the configuration of a solution that satisfies the preferences of a single user. The concept of \emph{multi-configuration} is based on the idea of configuring a set of configurations. Such a functionality is relevant in scenarios such as the configuration of personalized exams, the configuration of project teams, and the configuration of different trips for individual members of a tourist group (e.g., when visiting a specific city). In this paper, we exemplify the application of multi-configuration for generating individualized exams. We also provide a constraint solver performance analysis which helps to gain some insights into corresponding performance issues.
\end{abstract}

%%
%% Keywords. The author(s) should pick words that accurately describe
%% the work being presented. Separate the keywords with commas.

%%
%% This command processes the author and affiliation and title
%% information and builds the first part of the formatted document.
\maketitle

\section{Introduction}\label{introduction}

%TBD: Motivate the relevance of the concept of multi configuration, discuss different application scenarios, discuss related work, discuss major contributions of the paper, explain the organization of the paper.

%TBD: include further references \cite{Felfernigetal2014}...

Configuration is the process of assembling basic components into a complex product while taking into account a set of constraints \cite{AIcom1997, ConstraintProgrammingHandbookJunker2006, LHvam2008, Felfernigetal2014}. Most existing configurators are based on the assumption that a solution (configuration) is developed for a single user. On the contrary, group-based configuration \cite{ConfWS16,VietMan2022} focuses on the configuration of a solution for a group of users. Such a configuration must satisfy the preferences of each individual user as much as possible \cite{ConfWS14,AtasPsychologyAwarePreferenceConstruction}. Group-based configuration can be further extended to allow the configuration of a set of solutions based on the preferences of one or multiple users. Such a \emph{multi-configuration problem} \cite{ConfWS21} includes a set of constraints specifying restrictions with regard to (1) the combination of multiple solutions and (2) properties of a specific solution. Scenarios including such configuration sets may benefit from multi-configuration. Related example scenarios are the following.

 \emph{Multi-exam configuration}.
    individual exams are configured for each student, where related constraints are specified by instructors and possibly also students. A configurator can support instructors during the exam preparation phase and helps in the prevention of cheating through the generation of individualized exams \cite{ConfWS21exam}.

    \emph{Project team assignment}.
    persons are assigned to teams such that each team has the expertise to successfully complete the corresponding project (assigned to the team). In this context, fairness aspects can play a role, for example, each team should have at least similar chances to complete a project within pre-defined time limits \cite{ConfWS21}.

    \emph{Generation of test cases}.
    The automated generation of test cases can be considered as a multi-configuration problem, where input values need to be generated in such a way that given   coverage criteria are fulfilled \cite{ISSTA98}.

    \emph{Holiday planning}.
    Members of tourist groups often do not share the same interests during excursions \cite{AVI2004}, i.e., which sightseeing destinations to visit. Therefore, a configurator could configure different trips for subgroups of tourists based on their preferences.

    \emph{Configuration space learning}.
    Many (software) systems (e.g., operating systems) offer a high degree of configurability. In this context, it is difficult to find the optimal configuration settings \cite{PEREIRA2021111044} also due to the fact that it is impossible to test all possible settings to find the optimal one. Multi-configuration can support the identification of test configurations that help to learn dependencies between configuration parameters.

In the context of \emph{multi-exam configuration}, we have built a software library that helps to configure exams. Example inputs are the number of examinees, a pool of questions, and a set of constraints specifying preferences of instructors and examinees. The outcome is a set of questions for each examinee. Constraint solving in our implementation is based on the \textsc{Choco} constraint solver.\footnote{https://choco-solver.org/}

In this paper, we show how the problem of multi-exam configuration can be represented as a constraint satisfaction problem (CSP) \cite{handbookconstraintprogamming2006}. We exemplify different types of constraints supported by our configurator and also show how the preferences of students can (potentially) be taken into account. In order to analyze constraint solver performance, we evaluate the runtime performance of an open source constraint solver (\textsc{Choco}) on the basis of a typical real-world exam configuration scenario.

%In order to analyze constraint solver performance, we analyze the runtime performance of an open source constraint solver depending on the integration of different constraint types. Furthermore, we analyze the solver performance on the basis of a typical real-world exam configuration scenario.

The remainder of this paper is organized as follows. In Section \ref{workingexample}, we introduce a definition of a multi-configuration task and provide an example from the domain of multi-exam configuration. In this context, we also introduce and exemplify different constraint types. Thereafter, in Section \ref{evaluation}, we evaluate the performance of \textsc{Choco} when solving multi-exam configuration tasks also including a performance analysis when solving a real-world configuration task. Threats to validity are discussed in Section \ref{threatstovalidity}. The paper is concluded with a discussion of open research issues in Section \ref{conclusions}.

\section{Working Example}\label{workingexample}

We now give a basic definition of a \emph{multi-configuration task} (see Definition 1) (see \cite{ConfWS21}) and show how multi-exam configuration tasks can be introduced correspondingly.

\vspace{0.15cm}

\textit{Definition 1.} A multi-configuration task can be defined as a tuple $(V, D, C)$ with $V = \bigcup \{v_{ij}\}$ is a set of finite domain variables ($v_{ij}$ is variable $j$ of configuration instance $i$), $D = \bigcup \{dom(v_{ij})\}$  a set of corresponding domain definitions, and $C = \{c_1, c_2, ..., c_v\}$ a set of constraints.

%\textit{Definition 1.} A multi-configuration task can be defined as a tuple $(V, D, REQ, C)$ with $V = \bigcup \{v_{ij}\}$ being a set of finite domain variables, where $v_{ij}$ is variable $j$ of configuration instance $i$, $D = \bigcup \{dom(v_{ij})\}$ being a set of corresponding domain definitions, $REQ = \bigcup \{r_1, r_2, ..., r_u\}$ being a set of constraints representing user requirements, and $C = \{c_1, c_2, ..., c_v\}$ being a set of constraints \cite{ConfWS21}. $REQ$ and $C$ both constrain how certain variable values can be combined with each other.

\vspace{0.15cm}

In the following, we will use this definition to introduce the task for multi-exam configuration and outline what constraint types are supported by our configurator. In this context, the set of constraints $C$ can be defined by users represented by \emph{instructors} and also \emph{students} (where this is intended).

\subsection{Multi-exam configuration} \label{multi-exam-conf}

Following Definition 1, a multi-exam configuration task can be defined as follows.

\begin{itemize}
    \item $V = \{q_{11}..q_{nm}, q_{11}.T_1..q_{11}.T_\pi, .., q_{nm}.T_1..q_{nm}.T_\pi\}$ where $q_{ij}$ is question $j$ of the exam of student $i$, $n$ is the number of exams (students), $m$ is the number of questions per exam, $q_{ij}.T_k$ denotes the value of the \textit{k-th} \emph{question property} of question $q_{ij}$, and $\pi$ represents the number of question properties per question (in our case, $\pi=6$). Our configurator supports the following question properties: 
    \begin{enumerate}
        \item \emph{topic} - topic of the question
        \item \emph{level} - difficulty level of the question
        \item \emph{min-duration} - minimum estimated time needed to answer the question
        \item \emph{max-duration} - maximum estimated time needed to answer the question
        \item \emph{type} - type of the question (e.g. single/multiple choice, assignment task, etc.)
        \item \emph{points} - maximum number of points rewarded for correct answers
    \end{enumerate}

    \item $D = \{dom(q_{11})..dom(q_{nm}), dom(q_{11}.T_1)..dom(q_{11}.T_6), \allowbreak.., dom(q_{nm}.T_1)..dom(q_{nm}.T_6)\}$ where $dom(q_{ij}) = \{1..\Omega\}$, with $\Omega$ being the total number of questions in the question pool, and $dom(q_{ij}.T_k)$ is a question property domain (one out of the following):
    \begin{enumerate}
        \item \emph{dom(topic)} = $\{1..\eta\}$ where $\eta$ is the number of defined question topics
        \item \emph{dom(level)} = $\{1..\mu\}$ where $\mu$ is the number of defined question complexity levels
        \item \emph{dom(min-duration)} = $\{1..\tau\}$ with $\tau$ indicating the maximum specifiable value
        \item \emph{dom(max-duration)} = $\{1..\kappa\}$ with $\kappa$ indicating the maximum specifiable value
        \item \emph{dom(type)} = $\{1..\theta\}$ where $\theta$ is the number of defined question types
        \item \emph{dom(points)} = $\{1..\phi\}$ where $\phi$ is the maximum amount of points
    \end{enumerate}

    %\item $REQ = \{r_1..r_u\}$ where $r_\alpha$ is the requirement identifier and $u$ the number of user requirements

    \item $C = \{c_1..c_v\}$ where $c_\beta$ is the constraint identifier and $v$ is the number of constraints
\end{itemize}

Importantly, depending on the question $1..\Omega$ assigned to a question variable $q_{ij}$, a set of corresponding question properties must hold, for example, if question $q_{11}=1$, corresponding restrictions such as $q_{11}=1 \rightarrow q_{11}.topic=A$ indicate the relevant question properties. In Subsection \ref{subsec:flexiblebounds}, we explain in which way we support this aspect in our configuration library. Furthermore, for each student-specific exam $i$, we need to include an $\text{\emph{alldifferent}}(q_{i1} .. q_{im})$ constraint to avoid situations where a questions is assigned to the same exam twice. In our implementation, this aspect is taken into account on the basis of set variables (see also Subsection \ref{subsec:flexiblebounds}).

\subsection{Instructor constraints}

Instructor constraints (defined by instructors) in $C$ restrict the set of questions that may or may not appear in exams. Each exam must fulfil all of these constraints. We distinguish between two types of related constraints: \emph{intra-exam} and \emph{inter-exam} constraints.

\subsubsection{Intra-exam constraints}
Intra-exam constraints restrict which questions are eligible for being part of an exam. Such constraints refer to each individual exam. A simple form of intra-exam constraints is to directly define a specific question property. For example, let us assume that up to now a course has covered only one (the first) topic ($A$). As a consequence, the instructor requires that only questions belonging to topic $A$ are part of the first exam (see Formula \ref{eq:1}).

\begin{equation} \label{eq:1}
    \forall q_{ij} \in V : q_{ij}.topic = A
\end{equation}

Furthermore, we might want to restrict the complexity level of questions. For example, assuming that four different question complexities exist, for the final exam the instructor would like to increase the overall exam complexity using an intra-exam constraint specifying that all questions of each exam must have a complexity level of at least 2 (see Formula \ref{eq:2}).

%Inequality and specifying lower/upper bounds (including/excluding) are also supported ($=, \neq, >, <, \geq, \leq$). For example, assuming four different question complexities exist, for the final exam the instructor would like to increase the overall difficulty of the exams. Hence, they define an intra-exam constraint such that all questions of each exam have the complexity level of at least 2 (see Formula \ref{eq:2}).

\begin{equation} \label{eq:2}
    \forall q_{ij} \in V : q_{ij}.level \geq 2
\end{equation}

Intra-exam constraints allow instructors to arbitrarily combine multiple constraints with logical operators. For example, since multiple choice questions can generally be answered rather quickly, an instructor could require that every multiple choice question is of at least difficulty level 3 (see Formula \ref{eq:3} where we assume question type 3 indicates multiple choice questions).

\begin{equation} \label{eq:3}
    \forall q_{ij} \in V : (q_{ij}.type = 3 \implies q_{ij}.level \geq 3)
\end{equation}

In many cases, instructors would like to be able to specify intra-exam constraints on a more granular level. It is possible to combine intra-exam constraints with a corresponding scope. Constraint scopes enable instructors to specify how many questions per exam need to satisfy a given constraint. To illustrate this aspect, we will continue our previous example (see Formula \ref{eq:1}). By the time the next topic (topic $B$) is covered in the course, the students will have a follow-up exam consisting of 10 questions. The instructor now wants to focus mainly on the new topic. Therefore, they specify constraints such that for each exam only 2 questions belong to topic $A$ and the remaining 8 to topic $B$ (see Formula \ref{eq:4} and \ref{eq:4.1}).

\begin{equation} \label{eq:4}
   \bigwedge\limits_{i=1}^{n(\#exams)} (|\{q_{ij} \in V : q_{ij}.topic = A\}| = 2)
\end{equation}

\begin{equation} \label{eq:4.1}
   \bigwedge\limits_{i=1}^{n(\#exams)} (|\{q_{ij} \in V : q_{ij}.topic = B\}| = 8)
\end{equation}

Constraint scopes also support lower and/or upper bounds, for example, the instructor would like to keep the follow-up exam rather simple. For this reason, between 5 and 10 questions of each exam should be  easy to solve, which is indicated by complexity level 1 (see Formula \ref{eq:5}).

\begin{equation} \label{eq:5}
   \bigwedge\limits_{i=1}^{n(\#exams)} (5 \leq |\{q_{ij} \in V : q_{ij}.level = 1\}| \leq 10)
\end{equation}

Instructors may also specify constraint scopes using percentages in order to describe which amount of questions per exam must satisfy the question property constraint. For example, only between 10 and 20 percent of questions per exam should be solvable in less than five minutes (see Formula \ref{eq:6}).

\begin{equation} \label{eq:6}
   \bigwedge\limits_{i=1}^{n(\#exams)} \left ( 0.10 \leq \frac{|\{q_{ij} \in V : q_{ij}.min\text{-}duration < 5\}|}{m}  \leq 0.20 \right )
\end{equation}

Intra-exam constraints also support aggregations. In the context of our evaluation settings, we support the functions \emph{sum}, \emph{average}, and \emph{distinct count}. For their application, see the examples in Formulas \ref{eq:7}--\ref{eq:10}.

%Intra-exam constraints also support the integration of aggregate functions. When defining such constraints,  instructors may specify an aggregation function with respect to one selected question property. This will compute the specified function on all question per exam. The result of this computation is compared to a specified value. This may be extended to also allow lower/upper bounds (including). The supported aggregator functions are \emph{sum}, \emph{average}, and \emph{distinct count}. To give a better understanding of how these functions work, we will show one example constraint for each of them.

\begin{enumerate}
    \item Example (\emph{sum}): The total amount of points per exam is 100 (see Formula \ref{eq:7}).
    \begin{equation} \label{eq:7}
         \bigwedge\limits_{i=1}^{n(\#exams)} \left( \sum_{j=1}^{m} q_{ij}.points = 100 \right)
    \end{equation}

    \item Example (\emph{average}): The average complexity level of each exam is between 2 and 3. (see Formula \ref{eq:8}).
    \begin{equation} \label{eq:8}
         \bigwedge\limits_{i=1}^{n(\#exams)} \left( 2 \leq \frac{\sum_{j=1}^{m} q_{ij}.level}{m} \leq 3 \right )
    \end{equation}

    \item  Example (\emph{distinct count}): Each exam consists of at least 3 different question topics (Formula \ref{eq:10}).
    \begin{equation} \label{eq:10}
         \bigwedge\limits_{i=1}^{n(\#exams)} \left( |\{q_{ij}.topic\}| \geq 3 \right )
    \end{equation}
\end{enumerate}

\subsubsection{Inter-exam constraints}

Similar to intra-exam constraints, inter-exam constraints restrict which questions may be part of exams. However, they constrain how often certain question or question properties may or may not appear in the entire exam configuration. Therefore, inter-exam constraints depend on all exams combined, instead of every exam individually. Such constraints count, for example, how many exams have at least one question that fulfills a given constraint. This sum can be lower and/or upper bounded. For example, we assume that a specific question $\xi$ is part of at least 5 exams but at most 10 (see Formula \ref{eq:13}).

\begin{equation} \label{eq:13}
         5 \leq |\{q_{ij} \in V : q_{ij} = \xi\}|  \leq 10
\end{equation}

%\begin{equation} \label{eq:13}
%    5 \leq \sum_{i=1}^{n(\#exams)} \left ( \begin{cases}
%        1 & , \textnormal{ if } |\{q_{ij} \in V : q_{ij} = \xi\}| \geq 1 \\
%        0 & , \textnormal{ else}\\
%    \end{cases} \right ) \leq 10
%\end{equation}

As a special case of inter-exam constraints, instructors can restrict the degree of question overlap across exams, i.e., the number of questions that exams have in common. This is especially useful to prevent the generation of identical or very similar exams. The degree of overlap can be lower and upper bounded in order to restrict the minimum and maximum amount of questions that pairs of exams may share. For example (see Formula \ref{eq:14}), the upper bound denotes that no pair of exams exists which shares more than 5 questions, whereas the lower bound states that every pair of exams must share at least 2 questions. This might be useful to create a sense of fairness among students but might lead to cheating. $e_\lambda$ represents a set of questions comprising all questions of exam $\lambda$.

\begin{equation} \label{eq:14}
    \bigwedge\limits_{i=1}^{n(\#exams)} \bigwedge\limits_{j=1}^{n(\#exams)} \left ( 2 \leq | e_i \cap e_j | \leq 5 \right ) (i\neq j)
\end{equation}

The amount of exam pairs to be constrained can be further decreased in the context of onsite exams where a pre-defined lecture hall's seating plan (chart) is specified. One goal in such scenarios is to prevent cheating of students positioned in a neighborhood which can be achieved on the basis of constraints avoiding question overlaps in the case of students located next to each other. 

\vspace{0.1 cm}

\begin{figure}[ht]
    \centering
    \includegraphics[width=6.0cm]{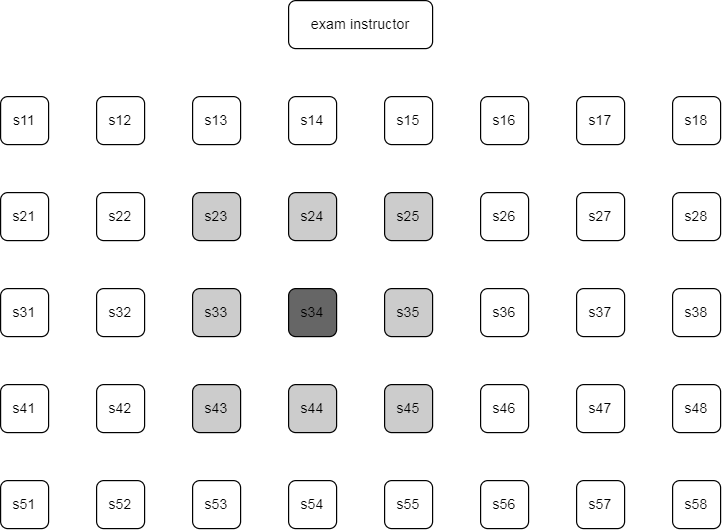}
    \caption{lecture hall seating: seat 34 and its neighbors.}
    \label{fig:seating_chart2}
\end{figure}

Given is the following lecture hall with 5 rows and 8 seats per row (see Figure \ref{fig:seating_chart2}). 

%\vspace{0.2cm}

The seats are labeled with the letter \emph{s} and two digits. The first digit represents its row (top to bottom) and the second digit its position in the row (left to right). We assume no neighboring exams may share even a single question (see Formula \ref{eq:15}), where $k$ is the number of neighbors of exam $i$, $e_i$ is the question set assigned to exam $i$ (student $i$), and $f(i, j, s)$ describes the set of questions of $i$'s \textit{j-th} neighboring exam according to seating chart $s$.

\begin{equation} \label{eq:15}
    \bigwedge\limits_{i=1}^{n(\#exams)} \bigwedge\limits_{j=1}^{k} \left ( | e_i \cap f(i, j, s) | = 0 \right )
\end{equation}

This allows identical exams in the configuration but never for students right next to each other. In order to provide a better understanding of how many constraints approximately need to be added per seat, we have highlighted seat $s34$ (dark gray) and its neighbors (light gray) as an example (see Figure \ref{fig:seating_chart2}). This particular seat has 8 neighbors requiring 8 constraints to be added.

\subsection{Student constraints}

Student constraints in $C$ can be specified by each student individually. They constrain only the student's  exam, no other exams are affected. Student constraints can only further narrow down instructor constraints. 

Example: The instructor specifies a constraint such that between $20$\% and $50$\% of the questions of each exam must belong to topic $A$ (see Formula \ref{eq:11}).

\begin{equation} \label{eq:11}
    \bigwedge\limits_{i=1}^{n(\#exams)} \left (0.20 \leq \frac{|\{q_{ij} \in V : q_{ij}.topic = A)\}|}{m}  \leq 0.50 \right )
\end{equation}

Student $a$ decides to restrict this constraint even further so that only a maximum of 25\% of questions of their exam belong to topic $A$ (see Formula \ref{eq:12}).
\begin{equation} \label{eq:12}
    \frac{|\{q_{aj} \in V : q_{aj}.topic = A)\}|}{m}  \leq 0.25
\end{equation}

\section{Evaluation}\label{evaluation}

We now present a performance analysis of our multi-configuration setting.\footnote{The source code for this performance analysis are available in \href{https://github.com/AIG-ist-tugraz/ExamMultiConf}{https://github.com/AIG-ist-tugraz/ExamMultiConf}}

\subsection{Real world example}\label{subsec:realworld}

We assume that (1) 450 students participate in an exam and (2) a  question pool of 45 questions individually associated with one out of four different topic areas (topics) is available. Each exam should consist of $m=10$ questions. We define the following constraints ($C$):

\begin{enumerate}
    \item Each exam should include questions related to at least 2 different topics. 
    \begin{displaymath}
         \bigwedge\limits_{i=1}^{n(\#exams)} \left( |\{q_{ij}.topic\}| \geq 2 \right )
    \end{displaymath}

    \item There is at most one multiple choice question.\footnote{Question type 3 was assumed to be multiple choice.}
    \begin{displaymath}
        \bigwedge\limits_{i=1}^{n(\#exams)} (|\{q_{ij} \in V : q_{ij}.type = 3\}| \leq 1)
   \end{displaymath}

    \item 10\% -- 20\% of the questions are assigned to complexity level 4 being the most complex one.
    \begin{displaymath}
        \bigwedge\limits_{i=1}^{n(\#exams)} \left ( 0.10 \leq \frac{|\{q_{ij} \in V : q_{ij}.level = 4\}|}{m} \leq 0.20 \right )
    \end{displaymath}

    \item Each exam should include questions resulting in $40$ points in total. 
    \begin{displaymath}
         \bigwedge\limits_{i=1}^{n(\#exams)} \left( \sum_{j=1}^{m} q_{ij}.points = 40 \right)
    \end{displaymath}

    \item Neighboring exams share at most 2 questions (assuming a hall $s$ with 22 rows and 21 seats/row).
    \begin{displaymath}
    \bigwedge\limits_{i=1}^{n(\#exams)} \bigwedge\limits_{j=1}^{k} \left (| e_i \cap f(i, j, s) | \leq 2 \right )
    \end{displaymath}
\end{enumerate}

Considering these specific requirements, the configurator yielded a solution in about \emph{one second}. %Reducing the question pool to 40 questions decreased the duration to slightly below one second. However, removing even more questions did not result in improved runtimes. The smallest question pool size for which the configurator found a solution was 24. A question pool with a size close to that causes a spike in duration. For instance, at around 30 question the solver takes 35 seconds on average. Moving closer to 40 question causes the runtime to drop to approximately one second. From there on the runtime steadily rises with increasing question pool size. The solver takes 1.2 seconds with a question pool size of 50 and 1.5 seconds with 60 question to find a solution. Naturally, the time it takes for the configurator to find solutions depends not only on the number of exams/questions but also on the combination of question properties. Therefore, using a different seed for randomly generating questions could show very different runtimes. 

\subsection{Dealing with flexible upper bounds}\label{subsec:flexiblebounds}

Instead of relying on a constant number of questions per student exam, it is also possible to support flexible lower and upper bounds. We support this aspect by utilizing \textsc{Choco} \emph{set variables}. Every student exam includes a set of questions. The domain of a set variable is (implicitly) defined by lower and upper bound sets. The lower bound is a set of questions that each exam must include, whereas the upper bound defines the maximum possible question set. In our case, the lower bound is  empty and the upper bound equals the question pool. %While solving, \textsc{Choco} creates a new set on the basis of the lower bound set and adds questions from the upper bound set to it. Each time a question is added the solver checks if any constraints are violated. %If that is not the case, the current set counts as a solution. Otherwise, \textsc{Choco} remembers this particular assignment and continues adding questions in a different order.

A varying number of questions per exam could trigger the need for further instructor constraints, for example, to restrict the allowed number of questions. Let us assume a defined question pool with $\Omega = 3$ questions ($\{1, 2, 3\})$ and $n = 2$ exams ($e_1$ and $e_2$) represented as \textsc{Choco} set variables (\emph{model} is a \textsc{Choco} \emph{model object}).

\begin{verbatim}
e1 = model.setVar(lb:{}, ub:{1, 2, 3})
e2 = model.setVar(lb:{}, ub:{1, 2, 3})
\end{verbatim}

Now, we want to specify that each exam $e_i$ (of student $i$) needs to include at least two and at most three questions. %(see Formula \ref{eq:9}).
In \textsc{Choco}, this constraint would be defined as follows.
\begin{verbatim}
e1.setCard(model.intVar(2, 3))
e1.setCard(model.intVar(2, 3))
\end{verbatim}

%\begin{equation} \label{eq:9}
%    \bigwedge\limits_{i=1}^{n(\#exams)} \left( 2 \leq |e_i| \leq 3 \right )
%\end{equation}

In this simplified setting, the possible solution sets for both, $e_1$ and $e_2$ are: $\{1, 2\}, \{1, 3\}, \{2, 3\}$, and $\{1, 2, 3\}.$

If we also want to define restrictions on allowed question properties, the solver needs to know the question properties of each individual question. For scalability reasons, we avoid to define question/property relationships on the basis of constraints. Instead, we support a key-value data structure that allows the identification of question properties on the basis of the corresponding question $ID \in \{1..\Omega\}$. Given such a structure, we are now able to define constraints referring to question properties, for example: in each exam, the number of questions of topic $A$ is exactly 2. 

In \textsc{Choco}, no related built-in constraints exist. We have defined a custom constraint by extending the \emph{Propagator} class and implemented the two required methods \emph{propagate} and \emph{isEntailed}. The former is called in each iteration of the solving process. It tries to find solutions by counting the number of questions that belong to the specified topic. If the current branch of the solving process cannot satisfy the constraint, a contradiction is indicated. When taking into account this constraint, the possible solutions for $e_1$ and $e_2$ are $\{1, 2, 3\}$ and $\{1, 3\}$.

\subsection{Evaluation with synthesized data}

We have also evaluated the solver performance with synthesized multi-exam configuration tasks along the dimensions of \emph{number of questions} and \emph{number of exams}. Each task utilizes the same 5 constraints as discussed in Subsection \ref{subsec:realworld}. We choose the lecture hall size depending on the amount of exams $n$, using the formula $\lceil\sqrt{n}\rceil$, since this is a fairly simple way to assure that all students will fit in the lecture hall and to keep a good ratio between rows and seats per row. The results of this performance evaluation are summarized in Table \ref{tab:3} showing acceptable runtime performances in the context of typical exam settings as well as extreme cases of around $1000$ questions and up to $1000$ students inducing solver runtimes up to nearly $3$ minutes. Notice that a smaller question pool size does not always result in faster runtime.

\begin{table*}[h]
\caption{Constraint solver (configurator) performance based on synthesized settings differing in terms of number of questions and number of student-specific exams using the constraints introduced in Subsection 3.1. In this context, $s$=seconds and $m$=minutes. Cells without unit of measurement represent runtimes in $milliseconds$.}
\label{tab:3}
\resizebox{\textwidth}{!}{
\begin{tabular}{ccccccccccccc}
\multicolumn{2}{c}{}                                                                                 & \multicolumn{11}{c}{\cellcolor[HTML]{BFBFBF}\textbf{Exams}}                                                                                                                                                                                                                                                                                                                                                                         \\
\multicolumn{2}{c}{\multirow{-2}{*}{}}                                                               & \cellcolor[HTML]{BFBFBF}\textbf{1} & \cellcolor[HTML]{BFBFBF}\textbf{5} & \cellcolor[HTML]{BFBFBF}\textbf{10} & \cellcolor[HTML]{BFBFBF}\textbf{25} & \cellcolor[HTML]{BFBFBF}\textbf{50} & \cellcolor[HTML]{BFBFBF}\textbf{75} & \cellcolor[HTML]{BFBFBF}\textbf{100} & \cellcolor[HTML]{BFBFBF}\textbf{250} & \cellcolor[HTML]{BFBFBF}\textbf{500} & \cellcolor[HTML]{BFBFBF}\textbf{750} & \cellcolor[HTML]{BFBFBF}\textbf{1000} \\
\rowcolor[HTML]{EFEFEF} 
\cellcolor[HTML]{BFBFBF}                                     & \cellcolor[HTML]{BFBFBF}\textbf{25}   & 14,33                              & 328,33                             & 397,67                              & 349,00                              & 425,33                              & 475,67                              & 516,00                               & 1183,33                              & 3340,33                              & 6,30s                                & 10,10s                                \\
\cellcolor[HTML]{BFBFBF}                                     & \cellcolor[HTML]{BFBFBF}\textbf{50}   & 21,67                              & 34,00                              & 80,33                               & 100,67                              & 143,33                              & 153,00                              & 183,33                               & 448,67                               & 1231,33                              & 2572,33                              & 3769,00                               \\
\rowcolor[HTML]{EFEFEF} 
\cellcolor[HTML]{BFBFBF}                                     & \cellcolor[HTML]{BFBFBF}\textbf{75}   & 26,67                              & 46,33                              & 63,00                               & 88,33                               & 155,33                              & 221,33                              & 234,00                               & 686,67                               & 1689,00                              & 3387,00                              & 5,66s                                 \\
\cellcolor[HTML]{BFBFBF}                                     & \cellcolor[HTML]{BFBFBF}\textbf{100}  & 23,67                              & 55,33                              & 80,33                               & 161,00                              & 172,67                              & 318,00                              & 314,00                               & 849,33                               & 2345,00                              & 4899,00                              & 7,36s                                 \\
\rowcolor[HTML]{EFEFEF} 
\cellcolor[HTML]{BFBFBF}                                     & \cellcolor[HTML]{BFBFBF}\textbf{150}  & 36,00                              & 93,67                              & 102,67                              & 193,67                              & 301,00                              & 462,00                              & 485,00                               & 1163,33                              & 3751,00                              & 7,08s                                & 12,80s                                \\
\cellcolor[HTML]{BFBFBF}                                     & \cellcolor[HTML]{BFBFBF}\textbf{250}  & 109,00                             & 146,33                             & 185,33                              & 383,33                              & 562,67                              & 725,33                              & 934,33                               & 2540,67                              & 7,21s                                & 14,38s                               & 20,98s                                \\
\rowcolor[HTML]{EFEFEF} 
\cellcolor[HTML]{BFBFBF}                                     & \cellcolor[HTML]{BFBFBF}\textbf{500}  & 114,00                             & 269,33                             & 407,33                              & 878,00                              & 1638,33                             & 2174,33                             & 3020,33                              & 7,64s                                & 19,27s                               & 35,75s                               & 55,30s                                \\
\cellcolor[HTML]{BFBFBF}                                     & \cellcolor[HTML]{BFBFBF}\textbf{750}  & 168,33                             & 415,67                             & 719,67                              & 1677,33                             & 3163,33                             & 5,02s                               & 5,09s                                & 15,92s                               & 33,41s                               & 1,14m                                & 1,50m                                 \\
\rowcolor[HTML]{EFEFEF} 
\multirow{-9}{*}{\cellcolor[HTML]{BFBFBF}\textbf{\rotatebox{90}{Questions}}} & \cellcolor[HTML]{BFBFBF}\textbf{1000} & 250,33                             & 651,67                             & 1052,67                             & 2942,67                             & 5,01s                               & 8,08s                               & 9,04s                                & 27,18s                               & 1,06m                                & 1,81m                                & 2,67m                                
\end{tabular}}
\end{table*}

\section{Threats to Validity}\label{threatstovalidity}
In the context of the reported evaluation, we have applied the standard settings of the used constraint solver. A major topic of further work is to further improve runtime performance on the basis of different approaches supporting the learning of solver search heuristics (see, e.g., \cite{Uta2022}). Fairness is a crucial aspect to be taken into account when it comes to the automated generation of exams. In this work, we have taken this aspect into account a.o. on the basis exam-specific criteria regarding the percentage of to-be-included questions that are related to a specific complexity level. For future work, we plan to further refine this aspect, for example, on the basis of optimization functions that help to balance the complexity of individual exams on a more fine-grained level. Finally, in real-world settings, we often have to deal with situations where a given set of constraints is inconsistent, i.e., no solution could be identified. In our future work, we will integrate corresponding repair concepts which will help users to find ways out from the so-called \emph{no solution could be found} dilemma. Such approaches can be based o.a. on model-based diagnosis \cite{RReiter1987}.

\section{Conclusions}\label{conclusions}

In this paper, we have introduced multi-configuration as a useful approach in scenarios requiring solution set configuration, for example, exam configuration and project team configuration. In the context of multi-exam configuration, we have shown a corresponding configuration task representation as a constraint satisfaction problem. We have evaluated the performance of the proposed approach on the basis of an example real-world configuration task as well as  a collection of synthesized configuration tasks (differing in terms of the number of pre-defined questions and the number of "to be generated" exams). Our future work will include the integration of further concepts supporting solver performance optimization. Furthermore, we will include features, for example, in terms of optimization functions, that help to take into account aspects such as fairness in a more explicit fashion. Finally, we plan to include concepts that will allow us to take into account historical data, for example, when generating a set of "new" exams, the frequency of questions already "used" in previous exams should be taken into account in order to avoid situations where specific questions are posed too often.

\bibliography{sample-ceur}

\end{document}